\documentclass{article}

\usepackage{arxiv}

\usepackage[utf8]{inputenc} % allow utf-8 input
\usepackage[T1]{fontenc}    % use 8-bit T1 fonts
\usepackage{hyperref}       % hyperlinks
\usepackage{url}            % simple URL typesetting
\usepackage{booktabs}       % professional-quality tables
\usepackage{amsfonts}       % blackboard math symbols
\usepackage{nicefrac}       % compact symbols for 1/2, etc.
\usepackage{microtype}      % microtypography
\usepackage{lipsum}		% Can be removed after putting your text content

\usepackage{graphicx}
\usepackage{caption}
\usepackage{float}
\usepackage{subcaption}
\usepackage{amsmath}

\renewcommand{\headeright}{}
\renewcommand{\undertitle}{Draft}

\title{\emph{batchboost}: regularization for stabilizing training with
	resistance to underfitting \& overfitting}

\author{
  Maciej A.~Czyzewski\\
  Institute of Computing Science\\
  Poznan University of Technology\\
  Piotrowo 2, 60-965 Poznan, Poland\\
  \texttt{maciejanthonyczyzewski@gmail.com} \\
}

\begin{document}
\maketitle

% BC learning: https://arxiv.org/pdf/1711.10284.pdf
% EfficientNet: https://arxiv.org/pdf/1905.11946.pdf
% Mixup: https://arxiv.org/pdf/1710.09412.pdf
% SamplePairing: https://arxiv.org/pdf/1801.02929.pdf
% ShakeDrop: https://arxiv.org/pdf/1802.02375.pdf
% ShakeShake: https://arxiv.org/pdf/1705.07485.pdf

\begin{abstract}
	Overfitting \& underfitting and stable training are an important challenges in
	machine learning.
	Current approaches for these issues are \emph{mixup}\cite{zhang2017mixup},
	\emph{SamplePairing}\cite{inoue2018data}
	and \emph{BC learning}\cite{tokozume2018between}.
	In our work, we state the hypothesis that mixing many images together can be more
	effective than just two.
	\emph{batchboost} pipeline has three stages:
	(a) pairing: method of selecting two samples.
	(b) mixing: how to create a new one from two samples.
	(c) feeding: combining mixed samples with new ones from dataset into batch (with ratio $\gamma$).
	Note that sample that appears in our batch propagates with
	subsequent iterations with less and less importance until the end of training.
	Pairing stage calculates the error per sample, sorts the samples and pairs
	with strategy: hardest with easiest one, than mixing stage merges two samples
	using \emph{mixup}, $x_1 + (1-\lambda)x_2$. Finally, feeding stage combines
	new samples with mixed by ratio 1:1. 
	\emph{batchboost} has 0.5-3\% better accuracy than the current
	state-of-the-art \emph{mixup} regularization on
	CIFAR-10\cite{krizhevsky2009learning} \&
	Fashion-MNIST\cite{xiao2017}.
	Our method is slightly better than SamplePairing technique
	on small datasets (up to 5\%).
	\emph{batchboost} provides stable training on not tuned parameters (like weight
	decay), thus its a good method to test performance of different architectures.
	Source code is at: \url{https://github.com/maciejczyzewski/batchboost}
\end{abstract}


\section{Introduction}
\label{sec:introduction}

\begin{figure}[H]
  \centering
\begin{minipage}{.3\textwidth}
  \hspace{-0.65cm}
  \includegraphics[totalheight=5.6cm]{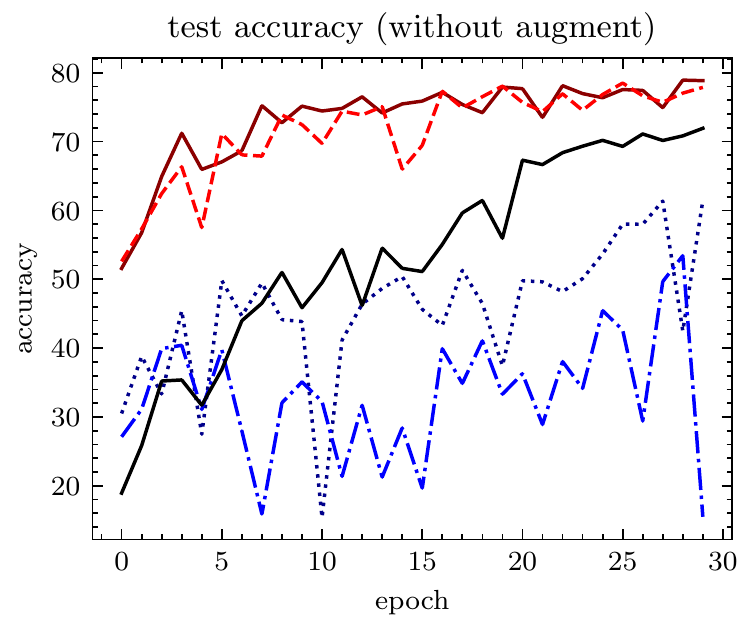}
\end{minipage}
\begin{minipage}{.65\textwidth}\vspace{-0.00cm}\hspace{0.865cm}
  \includegraphics[totalheight=5.6cm]{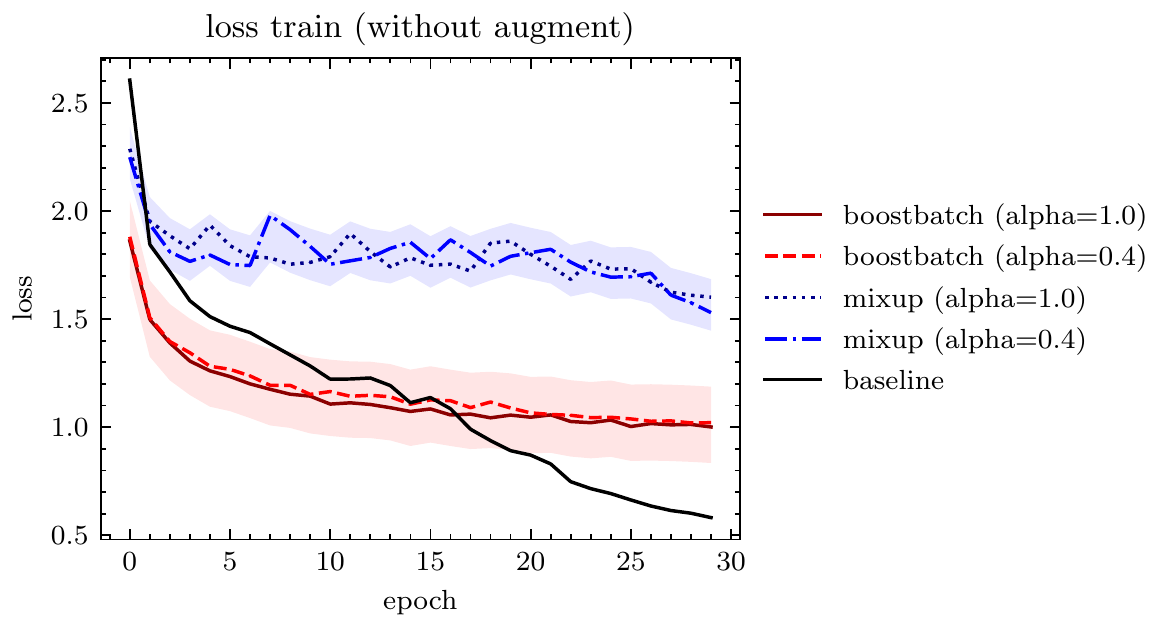}
\end{minipage}%
\caption{Evaluation on \emph{CIFAR-10}, for \emph{EfficientNet-b0} and
	\emph{SGD(weight-decay=10e-4, lr=0.1)} (as
	recommended in the \emph{mixup} research), same parameters for each model.
	As a result, the models behave differently, although they differ only in the
	method of constructing the batch.}
\label{fig:under}
\end{figure}

In order to improve test errors, regularization methods which are processes to
introduce additional information to DNN have been proposed\cite{miyato2018virtual}. Widely
used regularization methods include \emph{data augmentation}, \emph{stochastic
	gradient descent} (SGD) \cite{zhang2016understanding}, \emph{weight decay}
\cite{krogh1992simple}, \emph{batch normalization} (BN) \cite{ioffe2015batch},
\emph{label
	smoothing}\cite{szegedy2016rethinking} and \emph{mixup}\cite{zhang2017mixup}.
Our idea comes from \emph{mixup} flaws. In a nutshell, \emph{mixup} constructs
virtual training example from two samples. In term of batch construction, it
simply gets some random samples from dataset and randomly mix together.
The overlapping example of many samples (more than two) has not been considered
in previous work. Probably because the imposition of 3 examples significantly affects the model leading to underfitting.
It turned out that in many tasks, linear mixing (like \emph{BC learning} or
\emph{mixup}) leads to underfitting (figure \ref{fig:under}). Therefore, these methods are not applicable as universal tools.

\textbf{Contribution} Our work shows that the imposition of many examples in
subsequent iterations (which are slowly suppressed by new overlays) can improve efficiency, but most importantly it ensures stability of training and resistance to attacks.
However, it must be done wisely: that's why we implemented two basic mechanisms:
\begin{itemize}
\item (a) new information is provided gradually, thus \emph{half-batch} adds
new examples without mixing
\item (b) mixing is carried out according to some criterion, in our case it is the
best-the-worst strategy to mediate the error
\end{itemize}
The whole procedure is made in three steps to make it more understandable:
\begin{itemize}
\item (a) \emph{pairing}: a method for selecting two samples
\item (b) \emph{mixing}: how to create a new one from two samples
\item (c) \emph{feeding}: to the mixed samples it supplements the batch with new examples
from datasets
\end{itemize}
Note that sample that appears in our batch propagates with
subsequent iterations with less and less importance until the end of training.
Source code with sample implementation and experiments to verify the results
we present here:

\begin{center}
\url{https://github.com/maciejczyzewski/batchboost}
\end{center}

To understand the effects of \emph{bootstrap}, we conduct a
thorough set of study experiments (Section \ref{sec:results}).

\section{Overview}
\label{sec:overview}

\begin{figure}[H]
  \centering
  \includegraphics[width=\linewidth]{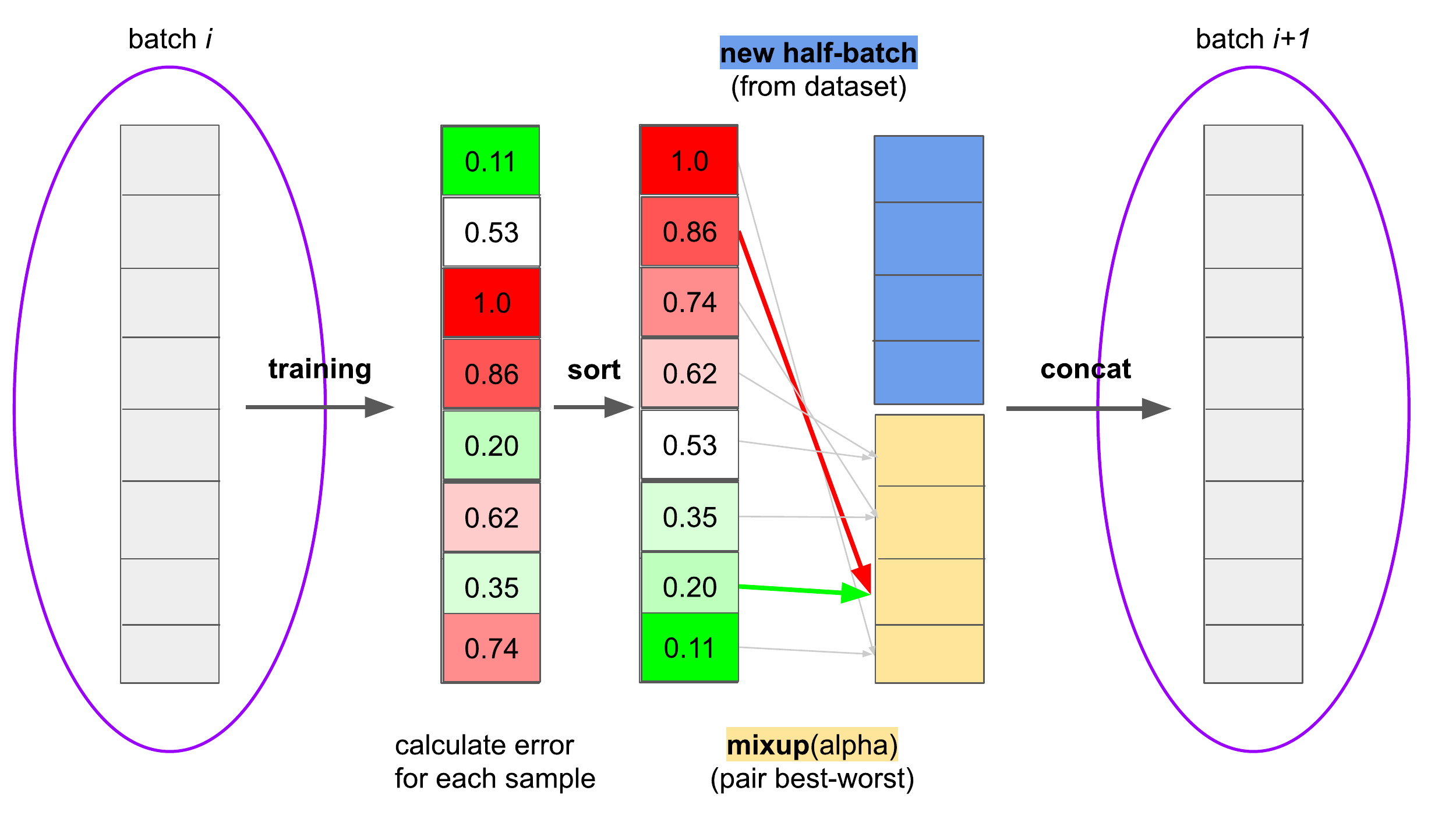}
  \caption{\emph{batchboost} presented in three phases: (a) pairing by sorting
	  error (b) mixing with \emph{mixup} (c) feeding: a mixed feed-batch and new
	  samples in half-batch by 1:1 ratio.}
  \label{fig:abstract}
\end{figure}

Batch as input for training is a combination of two different mini-batches:
\begin{itemize}
\item (a) \emph{half-batch}: new samples from dataset, classical augmentation is possible here
\item (b) \emph{feed-batch} (mixup): samples mixed together (in-order presented in
figure \ref{fig:abstract})
\end{itemize}

Parameter $\gamma$ means the ratio of the number of samples in half-batch to
feed-batch, in our work we have not considered other values than 1. However, we believe that this is an interesting topic for further research and discussion.

\subsection{Pairing Method}
\label{sec:pairing}

Combining many overlapping samples may have a negative impact on our optimizer
used in training.  In our implementation, it calculates the error for each
sample in batch.  Then it sorts this vector, and pairs samples by connecting the
easiest (smallest error) with the most difficult sample.  The goal of this
procedure is to create new artificial samples that are between classes, as
described in \emph{BC learning}.

However, in this case they are not random pairs, but those that 'require'
additional work. In this way, the learning process is more stable because there
are no cases when it mix only difficult with difficult or easy with easy (likely
is at the beginning or end of the learning process).
In our case, the error was calculated using L2 metric between one-hot labels and
the predictions (thus we analyzed \emph{batchboost} only on classification
problems like CIFAR-10\cite{krizhevsky2009learning} or
Fashion-MNIST\cite{xiao2017}). For other problems, there is probably
a need to change the metric/method of error calculation.
We were also thinking about using RL to pair samples. However, it turns out to
be a more complicated problem thus we leave it here for further discussion.

\subsection{Mixing Method}
\label{sec:mixing}

Selected two samples should be combined into one.
There are three methods for linearly mixing samples: \emph{SamplePairing},
\emph{Mixup}, \emph{BC Learning}. Due to the simplicity of implementation and
the highest scores, we used a mixup, which looks like this:
\begin{align*}
  \tilde{x} &= \lambda x_i + (1 - \lambda) x_j,\qquad \text{where~} x_i, x_j \text{~are~raw~input~vectors}\\
  \tilde{y} &= \lambda y_i + (1 - \lambda) y_j,\qquad \text{where~} y_i, y_j \text{~are~one-hot~label~encodings}
\end{align*}
$(x_i, y_i)$ and $(x_j, y_j)$ are two examples drawn at random from our
training data, and $\lambda \in [0,1]$.
Label for many samples was averaged over the last 2 labels (due to small differences in results, and large tradeof in memory).

Why it works?
The good explanation is provided in BC learning research, that images and sound
can be represented as waves. Mixing is an interpolation that human don't
understand but machine could interpret.
However, also a good explanation of this process is: that by training on
artificial samples, we supplement the training data by artificial examples between-classes
(visually, it fills space between clusters in UMAP/t-SNE visualization).
Thus, it generalizes problem more by aggressive cluster separation during
training (the clusters are moving away from each other, because model learns
artificial clusters made up by mixing).
Mixing multiple classes allows for more accurate separation (higher dimensions), however model starts to depart from original problem (new distribution) losing accuracy on test dataset.

The question is whether linear interpolation is good for all problems.
Probably the best solution would be to use a GAN for this purpose (two inputs +
noise to control). We tried to use the technique described in
SinGAN\cite{shaham2019singan} but it
failed in \emph{batchboost}.  It was unsuccessful due to the high cost of
maintaining such a structure.

\subsection{Continuous Feeding}
\label{sec:feeding}

The final stage is for 'feeding' new artificial samples on the model's input. In
the previous researches, considered were only cases with mixing two samples along
batch. \emph{batchboost} do this by adding new samples with $\gamma$ ratio to
mixed ones.
An interesting observation is that once we mix samples, they are in learning
process till end (at each batch continuously).
When applying a mixing it has only three options: (a) new sample with new sample
(b) new sample with previously mixed sample (c) previously mixed sample with
previously mixed sample. Pairing method cannot choose only one option for all samples
because of non-zero $\gamma$ ratio.

\begin{figure}[H]
  \hspace{0.5cm}
  \includegraphics[width=\linewidth]{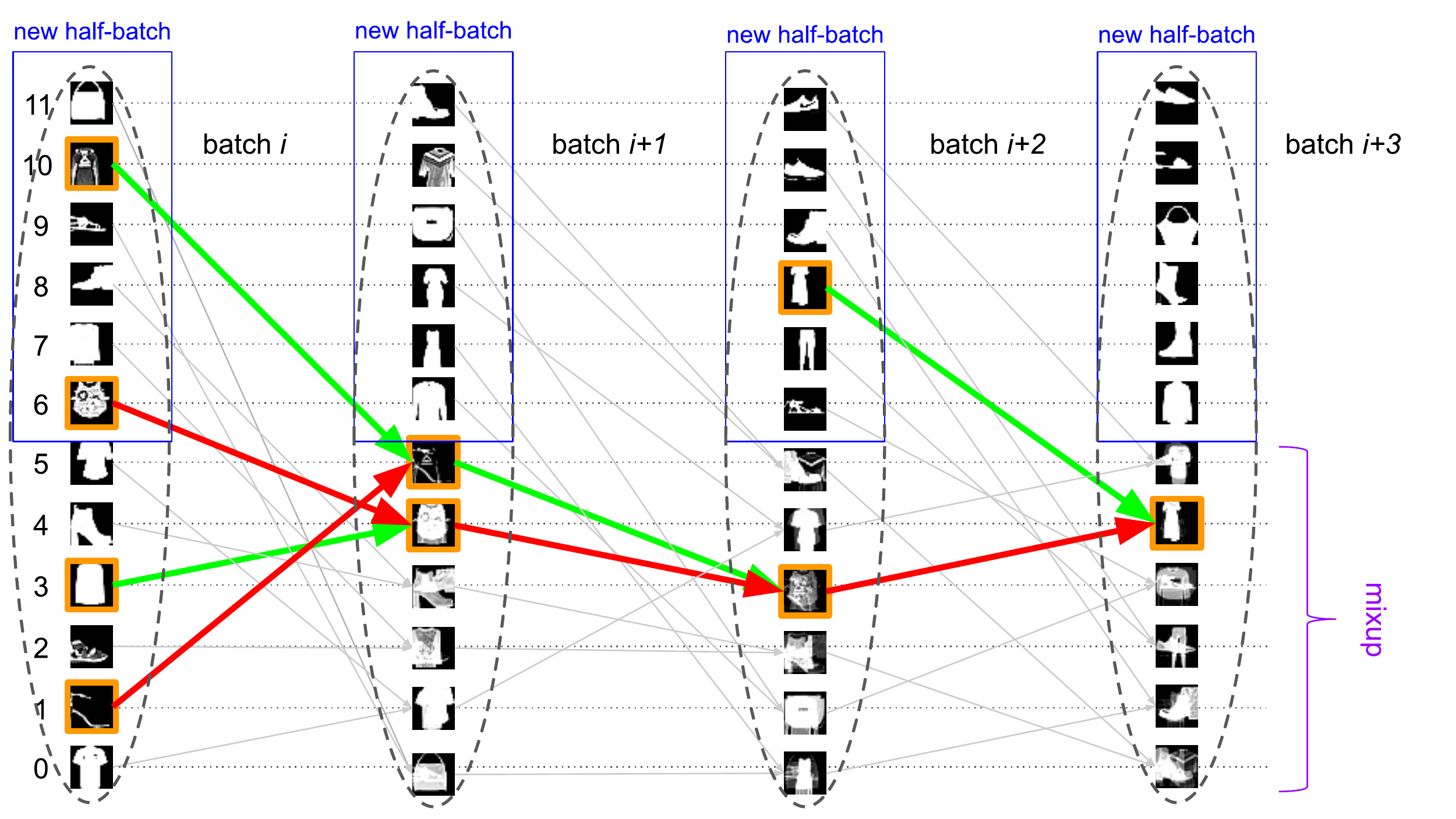}
  \caption{Orange squares indicates how information is propagated between
	  batches in the \emph{batchboost} method.}
  \label{fig:feeding}
\end{figure}

To maintain compatibility with the mixup
algorithm, it chooses new $\lambda$ when constructing the batch.
That is why past samples have less and less significance in training process,
until they disappear completely (figure \ref{fig:feeding}).

We found that for problems by nature not linear, for which the mixup did poorly,
it was caused by the fact that model learned at the time when very low/high
$\lambda$ was assigned (i.e. model learned on a single example, without mixing).
In \emph{batchboost} it doesn't look much better. However, \emph{half-batch}
contains new information, and \emph{feed-batch} has examples mixed not randomly but
by pairing method. With this clues, optimizer can slightly improve the direction of
optimization by better interpreting loss landscape.

\section{Results}
\label{sec:results}

We focused on the current state-of-the-art \emph{mixup}. The architecture we
used was \emph{EfficientNet-b0}\cite{tan2019efficientnet} and
\emph{ResNet100k}\cite{DBLP:journals/corr/HeZRS15} (having only 100k
parameters from DAWNBench\cite{coleman2017dawnbench}). The problems we've evolved are CIFAR-10 and
Fashion-MNIST.
We intend to update this work with more detailed comparisons and experiments,
test on different architectures and parameters. The most interesting
issue which requires additional research is artificial attacks.

\subsection{Underfitting \& Stabilizing Training}
\label{sec:under}

We described this problem in the (section \ref{sec:feeding}). The main factors
that stabilize training are: (a) the appropriate pairing of samples for mixing,
i.e. by error per sample (b) propagation of new information in \emph{half-batch}.

\begin{figure}[H]
  \centering
\begin{minipage}{.3\textwidth}
  \hspace{-0.65cm}
  \includegraphics[totalheight=5.6cm]{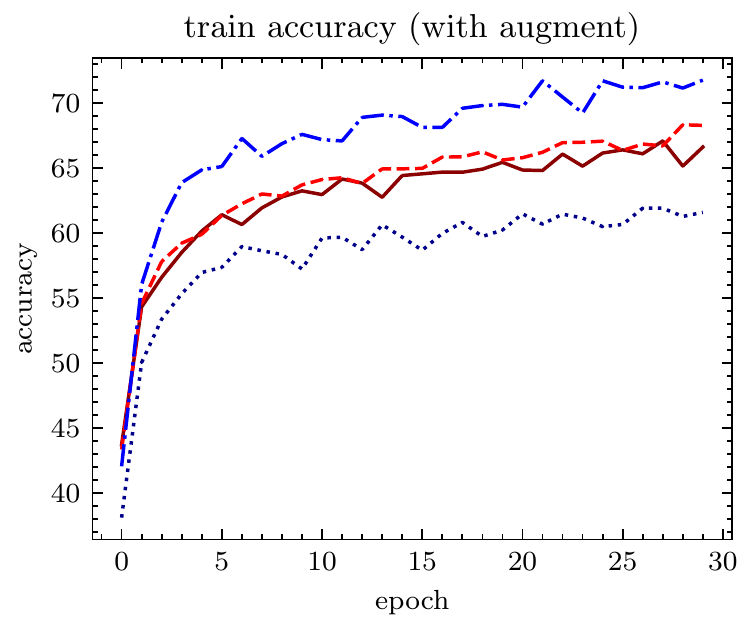}
\end{minipage}
\begin{minipage}{.65\textwidth}\vspace{-0.00cm}\hspace{0.865cm}
	\includegraphics[totalheight=5.6cm]{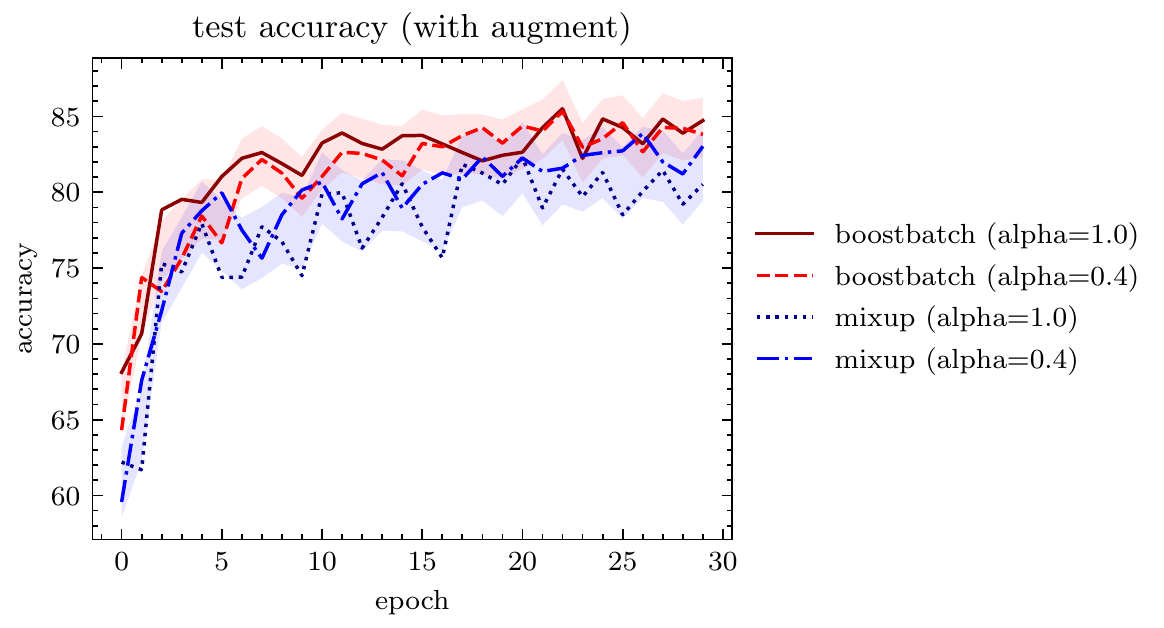}
\end{minipage}%
\caption{\emph{batchboost} is a new state-of-the-art because it is a slightly
	better than \emph{mixup} (here \emph{mixup} has been tuned for best
	parameters, \emph{batchboost} uses configuration from figure \ref{fig:under}).}
\label{fig:over}
\end{figure}

Another problem that \emph{mixup} often encounters is very unstable loss
landscape. Therefore, without a well-chosen weight decay, it cannot stabilize in
minimums. To solve this problem, we tune the optimizer parameters
for \emph{mixup}, after that it could achieve a similar result to
\emph{batchboost} (figure \ref{fig:over}).

\subsection{Overfitting (comparison to \emph{mixup})}
\label{sec:over}

The most important observation of this section is that \emph{batchboost} retains
the properties of the \emph{mixup} (similarly to \emph{SamplePairing} or
\emph{BC learning}). It protects against overfitting, having slightly better results.

The only difference is that the $\alpha$ coefficient from the original
\emph{mixup} is weakened.

\subsection{Accelerating Training \& Adversarial Attacks}
\label{sec:attacks}

\begin{figure}[H]
  \centering
  \includegraphics[width=10.5cm]{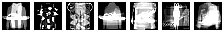}
  \caption{More than two samples have been mixed.}
  \label{fig:multipass}
\end{figure}

In the early stages, it learns faster than a classic \emph{mixup}.
The difference becomes significant when working on very small datasets, e.g.
medical challenges on Kaggle. In this work, we have limited \emph{Fashion-MNIST}
to 64 examples we compared to the classic model and \emph{SamplePairing}. The results were better by 5\$.
When the model perform well at small datasets, it means that training
generalizes problem. On (figure \ref{fig:multipass}) we present samples
generated during this process.

We tried to modify \emph{batchboost} to generate samples similar to those of
adversarial attacks (by uniformly mixing all samples backward with some Gaussian
noise) without any reasonable results.

\section{Conclusion}
\label{sec:conclusion}

Our method is easy to implement and can be used for any
model as an additional BlackBox at input.
It provides stability and slightly better results.
Using \emph{batchboost} is certainly more important in problems with small data sets.
Thanks to the property of avoiding underfitting for misconfigured parameters,
this is a good regularization method for people who want to compare two
architectures without parameter tuning.
Retains all properties of \emph{mixup}, \emph{SamplePairing} and \emph{BC learning}.

\bibliographystyle{unsrt}
\bibliography{references}

\end{document}